\documentclass{article} % For LaTeX2e
\usepackage{iclr2026_conference_arxiv,times}

\usepackage{amsmath,amsfonts,bm}

\def\eqref#1{equation~\ref{#1}}
\def\1{\bm{1}}

\DeclareMathAlphabet{\mathsfit}{\encodingdefault}{\sfdefault}{m}{sl}
\SetMathAlphabet{\mathsfit}{bold}{\encodingdefault}{\sfdefault}{bx}{n}

\usepackage[colorlinks=true,
            linkcolor=blue,
            citecolor=blue,
            urlcolor=blue]{hyperref}
\usepackage{xurl}            
\usepackage{url}
\usepackage{graphicx}
\usepackage{wrapfig}
\usepackage{booktabs} 
\usepackage{array}
\usepackage[most]{tcolorbox}
\usepackage{multirow}
\usepackage{enumitem}
\usepackage{makecell}

\title{On Open-Ended Information Seeking for Information Elicitation Agents}

\author{Victor De Lima \& Grace Hui Yang %\thanks{ Use footnote for providing further information about author (webpage, alternative address)---\emph{not} for acknowledging funding agencies.  Funding acknowledgements go at the end of the paper.} 
\\
InfoSense Lab, Georgetown University\\
}

\iclrarxivcopy

\begin{document}

\maketitle

\begin{abstract}

    Information elicitation is an open-ended information-seeking problem in which an interaction can unfold in many potentially valuable directions, requiring an elicitor to continually determine which information to pursue as new information emerges. In agentic elicitation, these decisions may be delegated to a foundation model, yet how model choice shapes the resulting information-seeking behavior remains understudied. We study how judgments about information value vary across LLMs and how these differences shape sequential information seeking. We first examine these judgments across 11 LLMs spanning multiple model families and parameter scales, using a shared set of information and elicitation objectives. We then develop a controlled elicitation simulation in which different models encounter the same information space and use the same selection rule, isolating these judgments from question generation and respondent behavior. Using this setting, we characterize the breadth--depth behavior that emerges from model-specific information-seeking preferences over the course of elicitation. We further examine how interaction history changes the evaluation and subsequent selection of prospective information. We test the robustness and boundaries of these findings through sensitivity analyses and ablations over the opportunities available to the elicitor, the response labels used to operationalize information-seeking preferences, the presence of interaction history, and whether redundancy is explicitly relevant to the assessment. The project code, data, and trajectory files are available at \url{https://github.com/infosenselab/open-elicitation}.

\end{abstract}

%% -- ################################################## %%
\section{Introduction}
%\vspace{-10pt}
%% -- ################################################## %%

Information elicitation is an open-ended information-seeking problem in which the objective is to maximize the \textit{value} of the facts uncovered through dialogue, a problem that arises in a wide range of real-world interactions, including recruiting, journalism, and oral history. It differs fundamentally from traditional information-seeking, which asks questions to distinguish among competing hypotheses by maximizing expected information gain \citep{feng_uncertainty_2024, chopra_feedback-aware_2025, lin_dual_2026}. Different lines of questioning may uncover entirely different sets of facts, from which equal or vastly different benefit can be derived \citep{DeLima2026}. When these decisions are delegated to an LLM-based elicitation agent, the choice of foundation model will shape what information is evaluated and prioritized. Yet how foundation model choice shapes this process over the course of an interaction remains understudied.

\begin{figure}
    
    \centering
    \includegraphics[height=5.2cm]{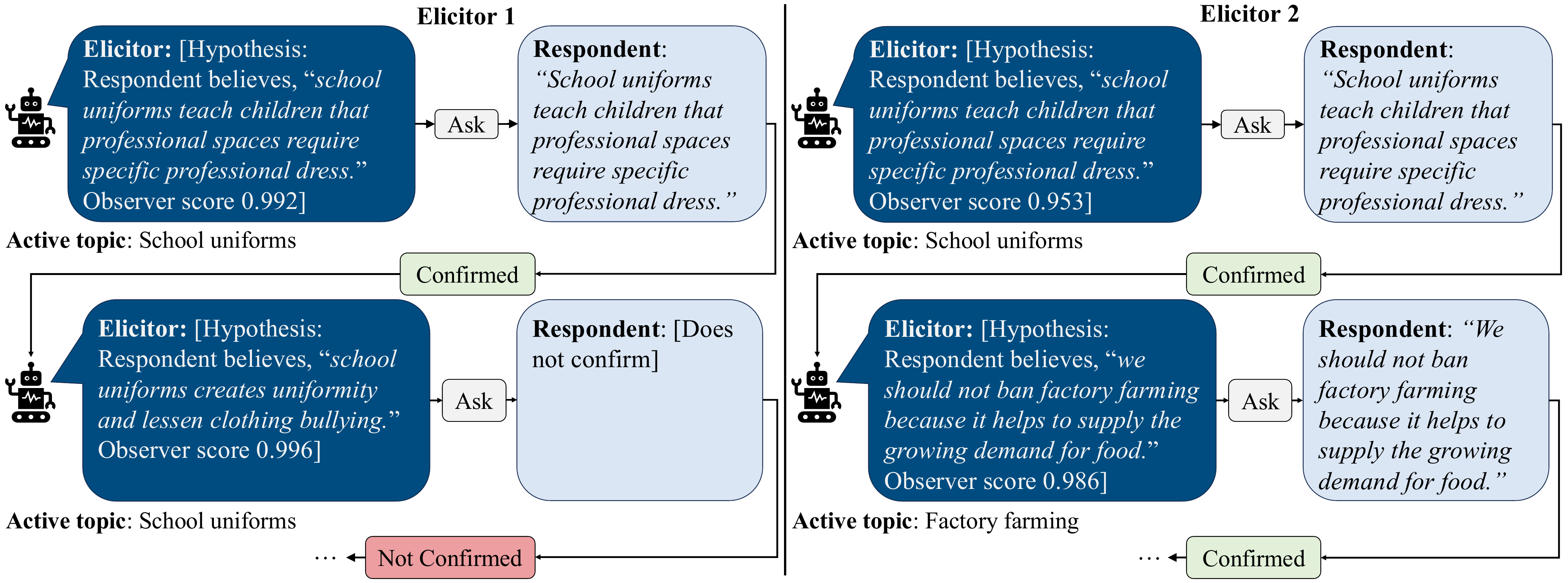}
    \vspace{-20pt}
    \caption{Illustration of the experimental elicitation setup and example trajectories.}
    \label{fig:elicitation_trajectories_illust}
    \vspace{-15pt}
\end{figure}

The open-ended nature of elicitation makes these prioritization decisions particularly consequential. Deciding what to select next, however, is not captured by information gain alone: two questions can convey the same amount of Shannon information \citep{shannon_mathematical_1948, yu_interactive_2020} while differing substantially in whether the information they reveal is worth pursuing. For example, asking whether a candidate is willing to relocate or prefers coffee to tea can provide one bit of Shannon information in either case, while differing dramatically in value to a recruiter. Classical Value of Information instead values information through its contribution to a downstream decision \citep{howard_information_1966, Grass2000}, but open-ended elicitation need not have a specified downstream decision. Nor can the problem generally be reduced to predefined information targets. Restricting an interaction to questions about where a candidate previously worked or whether they know Python reduces elicitation to a checklist that excludes unanticipated discoveries. Learning that a candidate unexpectedly ``\textit{worked at NASA},'' for example, may open a new line of inquiry about their mission, role, or challenges. As illustrated in Figure~\ref{fig:elicitation_trajectories_illust}, differences in how prospective information is prioritized can therefore redirect the subsequent course of elicitation and ultimately determine which regions of the information space are explored.

Studying how foundation model choice shapes information seeking presents several challenges. LLMs do not implement an explicit information-seeking policy from which their behavior can be readily anticipated: local preferences over prospective information may compound into trajectory-level patterns that only become apparent through sequential interaction. Moreover, information-seeking preferences vary with the objective, information already acquired, and opportunities available at each step, and therefore evolve over the course of an interaction. Lastly, information-seeking preferences interact with question generation, respondent behavior, and the information available at subsequent steps, making it difficult to isolate how the foundation model's preferences shape sequential elicitation.

To address these challenges, we make the following contributions:

\begin{itemize}[leftmargin=10pt, parsep=0pt]
    \vspace{-5pt}
    \item We systematically characterize information-seeking preferences across 11 open-weight instruction-tuned LLMs spanning multiple model families and parameter scales, multiple information-seeking objectives, and a shared information space.
    
    \item We introduce a sequential elicitation simulation that separates the evaluation of prospective information from question generation and respondent behavior, allowing model-dependent information-seeking preferences to be studied while holding other components fixed.

    \item We examine information-seeking preferences in sequential elicitation, studying how repeated candidate-level decisions shape the breadth and depth of elicitation trajectories and how interaction history changes the evaluation and subsequent selection of prospective information.

    \item We conduct extensive sensitivity analyses and ablations to establish the robustness and boundaries of these findings, varying the opportunities available to the elicitor, the response labels used to operationalize information-seeking preferences, the presence of interaction history, and whether redundancy is explicitly made relevant to the assessment.

\end{itemize}

%% -- ################################################## %%
\vspace{-10pt}
\section{Related Work}
\label{sec:related_work}
\vspace{-5pt}
%% -- ################################################## %%

A substantial line of work formulates information seeking as selecting observations or questions that reduce uncertainty about an underlying hypothesis, state, or parameter \citep{nelson_finding_2005}. This framing has been used to study human question asking \citep{testoni_efficiency_2023} and to guide language models through expected information gain \citep{mazzaccara_learning_2024, feng_uncertainty_2024, chopra_feedback-aware_2025} or the contribution of information-seeking steps toward reaching a target answer \citep{lee_prints_2026}. Related work in active sensing similarly values observations by how they change what an agent knows about its environment \citep{Veiga2023}. These approaches are well suited to settings such as diagnosis, troubleshooting, and identification, where information seeking is organized around uncertainty about an unknown target \citep{croskerry_clinical_2009, yazdani_models_2017}. In open-ended elicitation, there may instead be many possible discoveries worth pursuing rather than a predefined target whose uncertainty is to be resolved.

Decision-theoretic Value of Information (VoI) instead values information by its contribution to a downstream decision, distinguishing the amount of information acquired from its usefulness to a decision maker \citep{howard_information_1966,Grass2000}. VoI has been used to guide costly information acquisition \citep{bilgic_voila_2007,weiss_learning_2013} and extended to sequential decision making \citep{arumugam_value_2021}, human--AI decision making \citep{emmons_observation_2025,guo_value_2026}, and adaptive communication \citep{Dong2026}. The resulting value can depend strongly on the decision maker's preferences and utility function \citep{delquie_value_2008,abbas_value_2025}. This decision-theoretic framing does not directly cover open-ended elicitation, where information may be worth pursuing without a specified downstream decision.

Conversational information seeking studies information acquisition through multi-turn interaction \citep{dalton_conversational_2022, tang_reinforcement_2017}. Statistical dialogue management models such interaction as sequential decision making under uncertainty \citep{williams_partially_2007, young_pomdp-based_2013}, while mixed-initiative systems study when agents should answer, clarify, or proactively contribute information \citep{wu_inscit_2023,deng_proactive_2025, deng_towards_2024, wu_proactive_2019}. Other work focuses specifically on eliciting information from people \citep{bi_asking_2021, ren_learning_2021, li_learning_2022, kim_llm-as--interviewer_2024}, including follow-up questioning in interviews \citep{hu_designing_2024} and elicitation across domains such as academic interviews, journalism, judicial proceedings, and oral history \citep{DeLima2026}. Work on broader question spaces includes open-ended inquiry \citep{lin_dual_2026, lin_reward---line_2025, hamami_inquiry_2014}, curiosity-driven questions \citep{ko_inquisitive_2020}, and question salience for prioritizing questions by their contribution to understanding \citep{wu_which_2024}. Together, this work establishes settings in which many lines of inquiry are possible, but leaves open how LLMs prioritize among them and how those choices shape the course of elicitation.

Our work focuses on the information-seeking preferences that arise when LLMs prioritize among these possibilities. We isolate the evaluation of prospective information from question generation and respondent behavior, and study how these preferences shape sequential elicitation.

%% -- ################################################## %%
%\vspace{-10pt}
\section{Information-Seeking Preferences in LLMs}
\label{sec:information_seeking_preferences}
\vspace{-5pt}
%% -- ################################################## %%

%% ----------------------------------------------------- %%
%\vspace{-5pt}
\subsection{Operationalizing Information-Seeking Preferences}
\label{sec:information_seeking_score}
\vspace{-5pt}
%% ----------------------------------------------------- %%

We study how LLMs prioritize prospective information given an information-seeking objective and the information already acquired. Let $a$ denote an LLM, $h$ a candidate piece of information, $o$ the information-seeking objective, and $H_t$ the interaction history at step $t$. We represent the model's assessment with an information-seeking score $s_a(h\mid o,H_t)$, where higher scores indicate that model $a$ more strongly favors pursuing $h$ in the given context. We use these scores to characterize a model's \textit{information-seeking preferences}. This formulation allows preferences to depend on the model, objective, and interaction history without prescribing which considerations, such as relevance, novelty, specificity, or usefulness, the model should use in making the assessment.

For our empirical study, we operationalize each model's assessment using its probability of affirming that a candidate piece of information is worth pursuing for a specified objective and context. Prior work has shown that LLMs can evaluate the utility of information beyond semantic relevance, distinguishing information that is merely related to a query from information that is useful for addressing it \citep{zhang_are_2024, zhang_utility-focused_2025}. We leverage this capacity to study model judgments conditioned on a specified information-seeking objective and interaction history. Each model thereby induces its own information-seeking preferences. Because these probabilities need not be calibrated across models, we interpret them as model-specific information-seeking scores rather than directly comparable measurements of information value.

For each candidate--objective pair, the objective is incorporated into the system instruction and the argument is supplied as the user message using the model's native chat template (full prompt in Appendix~\ref{sec:appendix_prompts}). The model is asked whether the argument provides information worth pursuing for the specified objective and to choose one of three possible responses: a positive response $y_v$, a negative response $y_n$, and an uncertain response $y_u$. We include an uncertain option to reduce the tendency of the resulting scores to collapse toward extreme values \citep{cohen_i_2024} (see Appendix~\ref{sec:with_without_uncertain}). Rather than generating a response and treating the resulting discrete output as the model's assessment, we use a forward pass through the model to compute the probability assigned to each assessment label. This also avoids asking the model to generate a numerical score and treating that number as a direct measurement of its preference. Let $y=(y_1,\ldots,y_L)$ denote the token sequence corresponding to a candidate assessment, and let $x(h,o,H_t)$ denote the prompt presented to model $a$. We compute the teacher-forced log-probability of the complete assessment sequence as
$$
\ell_a(y\mid h,o,H_t)
=
\sum_{r=1}^{L}
\log P_a\!\left(
y_r
\mid
x(h,o,H_t),y_{<r}
\right).
$$
For assessments consisting of multiple tokens, this sums the token-level log-probabilities over the complete sequence. Exponentiating the resulting sum gives the probability assigned by the model to the candidate assessment. Let $y_v$ denote the affirmative assessment label. We define the information-seeking score used throughout our experiments as $s_a(h\mid o,H_t) = P_a(y_v\mid h,o,H_t)$.

%% ----------------------------------------------------- %%
%\vspace{-10pt}
\subsection{Characterizing Model Preferences}
\label{sec:information_seeking_characterization}
%% ----------------------------------------------------- %%

\begin{table}[t]
    %\vspace{-55pt}
    \centering
    \footnotesize
    \caption{Information-seeking scores ($P(\textit{Yes})$) across models to “\textit{A multi-party system could lead to more situations where a majority is not reached. This means that it could be harder to push policies through.}” Objectives are shown at the top, middle, and bottom of M14's ranking; SS denotes semantic similarity. All topics begin with ``we should.''}\label{tab:value_scores_args_observers}

    \begin{tabular}{
        m{5.2cm}
        >{\centering\arraybackslash}m{0.27cm}|
        >{\centering\arraybackslash}m{0.27cm}
        >{\centering\arraybackslash}m{0.27cm}
        >{\centering\arraybackslash}m{0.27cm}
        >{\centering\arraybackslash}m{0.27cm}
        >{\centering\arraybackslash}m{0.27cm}
        >{\centering\arraybackslash}m{0.27cm}
        >{\centering\arraybackslash}m{0.27cm}
        >{\centering\arraybackslash}m{0.27cm}
        >{\centering\arraybackslash}m{0.27cm}
        >{\centering\arraybackslash}m{0.27cm}
        >{\centering\arraybackslash}m{0.27cm}
    }
    \toprule
    Evaluation Topic $\downarrow$ Model $\rightarrow$ & SS & L3 & L8 & L70 & Q7 & Q14 & Q72 & M8 & M14 & M24 & P4 & G12 \\
    \midrule

    ...adopt a multi-party system (original)	& .38 & .97 &	.94 &	1.00 &	1.00 &	1.00 &	1.00 &	.86 &	.84 &	.99 &	.93 &	1.00\\

    \midrule

    \multicolumn{12}{l}{\textit{Positions 2-4}}\\

    ...oppose collectivism       	& .02 & .97 &	.82 &	1.00 &	.00 &	.00 &	.00 & 	.72 & 	.75 &	.84 &	.69 &	.99\\
    ...introduce compulsory voting	& .19 & .97 &	.86 &	1.00 &	.11 &	.00 &	.16 & 	.75 & 	.75 &	.68 &	.73 &	.84\\
    ...adopt libertarianism      	& .01 & .96 &	.79 &	1.00 &	.00 &	.00 &	.00 & 	.74 & 	.74 &	.94 &	.82 &	1.00\\

    \midrule
    
    \multicolumn{12}{l}{\textit{Positions 35-37}}\\
    ...ban missionary work               	& .01 & .81 &	.12 &	.00 &	.00 &	.00 &	.00 & 	.03 & 	.09 &	.03 &	.41 &	.00\\
    ...close Guantanamo Bay detention camp	                & .01 & .42 &	.21 &	.00 &	.00 &	.00 &	.00 & 	.02 & 	.08 &	.01 &	.54 &	.00\\
    ...abolish capital punishment        	& .01 & .43 &	.21 &	.00 &	.00 &	.00 &	.00 & 	.05 & 	.08 &	.01 &	.50 &	.00\\

    \midrule
    
    \multicolumn{12}{l}{\textit{Positions 69--71}}\\
    ...subsidize embryonic stem cell research          	& .01 & .46 &	.40 &	.00 &	.00 &	.00 &	.00 & 	.03 & 	.01 &	.01 &	.45 &	.00\\
    ...ban cosmetic surgery for minors       	& .01 & .08 &	.25 &	.00 &	.00 &	.00 &	.00 & 	.02 & 	.01 &	.04 &	.48 &	.00\\
    ...ban the use of child actors           	& .01 & .32 &	.20 &	.00 &	.00 &	.00 &	.00 & 	.01 & 	.01 &	.04 &	.33 &	.00\\
    
    \bottomrule
    \end{tabular}
    %\vspace{-15pt}
\end{table}

\begin{wraptable}[17]{r}{0.58\linewidth}
    \vspace{-25pt}
    \centering
    \footnotesize
    \caption{Agreement between models in their information-seeking preferences across objectives, with STSB semantic similarity (SS) included as a reference.}\label{tab:agreement_between_observers}

    \begin{tabular}{
        m{.2cm}
        >{\centering\arraybackslash}m{.2cm}
        >{\centering\arraybackslash}m{.2cm}
        >{\centering\arraybackslash}m{.2cm}
        >{\centering\arraybackslash}m{.2cm}
        >{\centering\arraybackslash}m{.2cm}
        >{\centering\arraybackslash}m{.2cm}
        >{\centering\arraybackslash}m{.2cm}
        >{\centering\arraybackslash}m{.2cm}
        >{\centering\arraybackslash}m{.2cm}
        >{\centering\arraybackslash}m{.2cm}
        >{\centering\arraybackslash}m{.2cm}
        >{\centering\arraybackslash}m{.35cm}
    }
    \toprule
     & SS & L3 & L8 & L70 & Q7 & Q14 & Q72 & M8 & M14 & M24 & P4 & G12 \\
    \midrule
	
        SS 	 & 1 & 0.23 & 0.16 & 0.29 & 0.25 & 0.28 & 0.21 & 0.24 & 0.32 & 0.32 & 0.14 & 0.34\\
        \midrule
    
        L3   &    & 1  &	.25     & .53   & .33   & .37   & .31	& .41 	& .53 	& .43 	& .14 	& .39   \\
        L8   &    &       & 1	   & .27    & .20   & .32   & .22	& .26 	& .29 	& .26 	& .22   & .24   \\
        L70  &    &       &		   & 1   & .37   & .39	& .47	& .50	& .60	& .51	& .13   & .49   \\
        Q7   &    &       & 		   &		& 1  & .28   & .33  & .31 	& .36 	& .34 	& .18   & .38	\\
        Q14  &    &       &         &		&		& 1	& .37  & .34 	& .47 	& .45 	& .22   & .42	\\
        Q72  &    &       &         &    	&		&		& 1	& .36	& .39	& .42	& .20	& .44  \\
        M8   &    &       &         &        &  		&		&		& 1 	& .59 	& .47 	& .15   & .42	\\
        M14  &    &       &         &        &      	&		&		&		& 1  & .60 	& .09   & .51	\\
        M24  &    &       &         &        &       &		&		&		&		& 1  & .09   & .53	\\
        P4   &    &       &         &    	&       &     	&		&		&		&		& 1  & .09	\\
        G12   &    &       &         &        &       &       &		&		&		&		&		& 1 \\

    \bottomrule
    \end{tabular}
\end{wraptable}

We evaluate 11 publicly available open-weight instruction-tuned LLMs, selected to provide variation both across model families and, where possible, across parameter scales within a family. We include Llama-3.2-3B-Instruct \citep{meta_llama_llama-32-3b-instruct_2024}, Llama-3.1-8B-Instruct \citep{meta_llama_llama-31-8b-instruct_2024}, Llama-3.3-70B-Instruct \citep{meta_platforms_inc_llama-33-70b-instruct_2024}, Qwen2.5-7B-Instruct, Qwen2.5-14B-Instruct, Qwen2.5-72B-Instruct \citep{yang_qwen2_2024}, Ministral-3-8B-Instruct-2512, Ministral-3-14B-Instruct-2512, and Mistral-Small-3.2-24B-Instruct-2506 \citep{liu_ministral_2026}. To broaden the comparison beyond these three families, we additionally include Phi-4-mini-instruct \citep{microsoft_phi-4-mini_2025} and gemma-4-12B-it \citep{gemma_team_gemma_2026}. In Tables~\ref{tab:agreement_between_observers} and~\ref{tab:value_scores_args_observers}, L, Q, M, P, and G denote the Llama, Qwen, Mistral/Ministral, Phi, and Gemma families, respectively, and the accompanying number denotes model size in billions of parameters. As a semantic-relatedness baseline, we use \textit{cross-encoder/stsb-roberta-large}~\citep{reimers_sentence-bert_2019}, which receives only the argument and evaluation topic.

We first characterize the information-seeking preferences induced by different LLMs before examining their consequences in sequential elicitation. We use the IBM-Rank-30k dataset~\citep{gretz_large-scale_2019}, which contains 30,497 short, human-written arguments collected in response to 71 researcher-selected controversial topics. For example, for the topic \textit{We should abandon the use of school uniforms}, one argument states that ``\textit{school uniforms do not allow children to emphasize their differences, denying individuality}.'' Each argument was subsequently evaluated by 10 annotators on whether they would recommend using the argument as-is in a speech supporting or contesting the topic. The dataset therefore provides a shared information space of naturally occurring arguments associated with known topics and varying substantially in their specificity and quality. To keep the evaluation computationally tractable, we randomly sample two arguments from each of the 71 topics, yielding 142 arguments. For each argument, we vary the objective $o$ across all 71 topics while keeping the candidate information $h$ unchanged. We use no interaction history ($H_t=\emptyset$), so that the objective and model are the only conditioning variables that vary. This produces $142\times71=10{,}082$ argument--objective evaluations per model. Because each evaluation is deterministic under a fixed model and prompt, we do not repeat evaluations across random seeds.

We instantiate the positive, negative, and uncertain response options $y_v$, $y_n$, and $y_u$ as \textit{Yes}, \textit{No}, and \textit{Uncertain}, respectively. We use \textit{Yes} and \textit{No} as the primary response labels because they provide simple, unambiguous expressions of positive and negative responses and are typically represented by a single token under the BPE-based tokenizers used by the models we study. We validate the robustness of this choice against alternative labels in Appendix~\ref{sec:label_trajectory_sensitivity}. With $H_t=\emptyset$ and $y_v=\textit{Yes}$, the information-seeking score becomes $s_a(h\mid o)=P_a(\textit{Yes}\mid h,o)$. We use $P(\textit{Yes})$ directly rather than renormalizing over \textit{Yes}, \textit{No}, and \textit{Uncertain}. Renormalization would instead measure the model's preference for \textit{Yes} conditional on its probability mass being restricted to the specified response set. 

Table~\ref{tab:agreement_between_observers} quantifies agreement in information-seeking preferences under identical candidate information and objectives. Agreement varies substantially across models, with most pairwise correlations ranging from .20 to .60 and P4 showing consistently lower agreement (.09--.22). Correlations with semantic similarity are low for all models (.14--.34), indicating that these preferences are not explained by semantic relatedness alone. Table~\ref{tab:value_scores_args_observers} shows that the variation nevertheless retains clear structure in a representative example. All models assign high information-seeking scores to the argument under its original, semantically aligned topic (.84--1.00), while substantial disagreement emerges for other objectives. For example, for \textit{We should oppose collectivism}, L3, L70, and G12 assign scores of .97, 1.00, and .99, whereas the Qwen models assign scores at or near zero. Taken together, these results show that the models induce substantially different, but not arbitrary, information-seeking preferences over the same information space.

%% -- ################################################## %%
%\vspace{-10pt}
\section{Elicitation Trajectory Simulation}
\label{sec:elicitation_trajectory_simulation}
\vspace{-5pt}
%% -- ################################################## %%

%% ----------------------------------------------------- %%
%\vspace{-5pt}
\subsection{Candidate Hypotheses and Transitions}
\label{sec:candidate_hypotheses}
\vspace{-5pt}
%% ----------------------------------------------------- %%

We simulate an interaction between an \textit{elicitor} and a \textit{respondent} using the IBM-Rank-30k information space, treating each argument as a hypothesis about a possible view of the respondent. At each step, the elicitor selects from $k$ candidate hypotheses, the respondent confirms or does not confirm the selected view, and the selection and response enter the interaction history. Repeating this process produces an elicitation trajectory. The dynamic is illustrated in Figure~\ref{fig:elicitation_trajectories_illust}. Let $C_t$ denote the set of $k$ candidate hypotheses presented to the elicitor at step $t$. Each hypothesis $h \in C_t$ is instantiated by an argument from the dataset, and $z(h)$ denotes the dataset topic associated with that argument. Before any hypothesis has been selected, no topic is active, and the initial candidate set is sampled across distinct topics. After the elicitor selects hypothesis $h_t$, its topic $z(h_t)$ becomes the \textit{active topic}, regardless of whether the respondent subsequently confirms the hypothesis. Subsequent candidate sets provide more opportunities to continue the active topic than to switch topics, while ensuring that alternative lines of inquiry remain available. Specifically, we draw the smallest possible majority of candidates from the active topic,
$k_{\mathrm{active}} = \left\lfloor \frac{k}{2} \right\rfloor + 1$, and draw the remaining ($k-k_{\mathrm{active}}$) candidates from distinct non-active topics. If the requested composition cannot be filled from these sources, the remaining positions are sampled from the pool of available hypotheses. Once a hypothesis has been selected, it is removed from future consideration. Unselected candidates remain available and may appear again at later steps. Because each selection determines the next active topic and therefore affects the composition of the next candidate set, the simulation is path dependent: $C_t \rightarrow h_t \rightarrow C_{t+1} \rightarrow h_{t+1} \rightarrow \cdots$.

%% ----------------------------------------------------- %%
\vspace{-5pt}
\subsection{Elicitor}
\vspace{-5pt}
%% ----------------------------------------------------- %%

We instantiate the elicitor as an information elicitation agent whose candidate-selection scores are provided by an LLM. For each candidate hypothesis $h\in C_t$, the prompt specifies both the overall objective of learning about the respondent's views and the candidate's topic $z(h)$. The information-seeking score is additionally conditioned on the previous hypotheses investigated about that same topic. That is, when scoring candidate $h$, the elicitor includes all previously selected hypotheses whose topic is $z(h)$, with each hypothesis followed by a \textit{[confirmed]} or \textit{[not confirmed]} label indicating the respondent's response. History concerning other topics is not included in that candidate's prompt.  Using the information-seeking score defined in Section~\ref{sec:information_seeking_score}, the elicitor scores each candidate and selects the candidate with the highest score, $h_t = \arg\max_{h\in C_t} s_a(h\mid o,H_t^{z(h)})$, where $H_t^{z(h)}$ denotes the interaction history associated with the candidate's topic. Ties between candidates with numerically equal maximum scores are broken randomly. Thus, the explicit selection rule is identical across elicitors; experimental conditions differ in the LLM used to assign information-seeking scores. Importantly, we do not generate the natural-language questions asked during the dialogue. Instead, when the elicitor selects a hypothesis, we treat that selection as choosing to ask about the view represented by the hypothesis and assume that an appropriate question is asked. The respondent then either confirms or does not confirm that view. In this way, the simulation preserves the sequential consequences of deciding what information to pursue while abstracting away variation due to natural-language question generation.

%% ----------------------------------------------------- %%
\vspace{-5pt}
\subsection{Respondent}
\label{sec:trajectories_respondent}
\vspace{-5pt}
%% ----------------------------------------------------- %%

Once $h_t$ has been selected, we independently sample whether the respondent confirms the corresponding view with fixed probability $p$. If confirmed, we treat this as learning that the respondent expresses the view represented by $h_t$, without making any judgment about whether its content is factually correct (factual correctness is a separate question, addressed by fact-checking and truth-seeking research). If unconfirmed, the simulation establishes only that the respondent did not confirm that view. The response is sampled independently of the hypothesis, topic, previous responses, and information-seeking score. This deliberately avoids making the resulting trajectories depend on a particular model of human response behavior. Both confirmed and unconfirmed outcomes are recorded in the interaction history and can therefore influence the information-seeking scores assigned to later hypotheses concerning the same topic. We also do not impose consistency constraints across the respondent's confirmed views. Although IBM-Rank-30k annotates whether arguments support or oppose their topics, enforcing consistency from these annotations would assume that views on a controversial topic reduce to a single position, even though a respondent may reasonably endorse arguments on both sides. For example, someone may agree that school uniforms reduce bullying while also agreeing that they limit individuality.

%% -- ################################################## %%
\vspace{-10pt}
\section{Experiments}
\label{sec:experiments}
\vspace{-5pt}
%% -- ################################################## %%

%% ----------------------------------------------------- %%
\subsection{Experimental Setup}
%% ----------------------------------------------------- %%

\begin{wraptable}[15]{r}{0.45\linewidth}
    \vspace{-70pt}
    \centering
    \footnotesize
    \caption{Breadth--depth behavior at $k=7$ (three-seed means). Sel./Cand.: selected/candidate WA; Max Rate: maximum-WA selection rate; Switch: topic-switch rate; Topics: distinct topics; Run: mean run length.}\label{tab:breadth_depth_model_set}

    \begin{tabular}{
        m{.4cm}
        >{\centering\arraybackslash}m{0.5cm}
        >{\centering\arraybackslash}m{0.5cm}
        >{\centering\arraybackslash}m{0.5cm}
        >{\centering\arraybackslash}m{0.5cm}
        >{\centering\arraybackslash}m{0.5cm}
        >{\centering\arraybackslash}m{0.5cm}
    }
    \toprule
     & Sel. WA & Cand. WA & Max Rate & Switch & Topics & Run \\
        \midrule

        Rand & .792 & .791 & .228 & .423 & 32.9 & 2.34\\

        \midrule
        
        L3	& .856	& .792	& .339	& .348	& 27.2 & 2.83\\
        L8	& .860	& .793	& .329	& .458	& 36.5 & 2.16\\
        L70	& .799	& .789	& .231	& .504	& 39.6 & 1.96\\
        Q7	& .847	& .791	& .315	& .551	& 43.8 & 1.79\\
        Q14	& .840	& .791	& .292	& .560	& 44.7 & 1.76\\
        Q72 & .841	& .790	& .286	& .525	& 42.0 & 1.88\\
        M8	& .837	& .790	& .290	& .413	& 34.0 & 2.54\\
        M14	& .818	& .792	& .260	& .526	& 42.7 & 1.88\\
        M24	& .818	& .791	& .258	& .484	& 39.0 & 2.04\\
        P4	& .861	& .790	& .351	& .295	& 23.4 & 3.35\\
        G12	& .814	& .791	& .241	& .342	& 26.2 & 2.89\\
        
    \bottomrule
    \end{tabular}
\end{wraptable}

We use the simulation introduced in Section~\ref{sec:elicitation_trajectory_simulation} to focus on two properties of model-dependent information-seeking preferences: (1) how these preferences shape the breadth and depth of elicitation trajectories and (2) how acquired information can change subsequent preferences and selections. Because the dataset contains arguments expressing different views on controversial topics, we set the elicitor's objective as ``\textit{Learn about an interview participant's views on common controversial topics}.'' Based on the candidate-set size analysis in Section~\ref{sec:available_opportunities}, we use $k=7$ unless otherwise specified. We fix the hypothesis success probability at $p=0.5$, giving each selected hypothesis an equal chance of being confirmed or unconfirmed. Because confirmation does not determine the opportunities available along a trajectory, this balances the two possible outcomes without favoring either. For each experimental condition, we simulate 100 trajectories of 100 elicitation steps across three random seeds ($42$, $43$, and $44$), yielding 30,000 elicitation steps in total. All experiments are run on three NVIDIA RTX A6000 GPUs. We include random selection as a trajectory baseline. We evaluate semantic similarity separately in Section~\ref{sec:information_seeking_characterization}, since there is no principled choice of text against which to compare candidates under a history-dependent assessment. We release the elicitation trajectory files generated for each model.

\begin{table}
    %\vspace{-25pt}
    \centering

    \caption{Effect of candidate-set size $k$ on elicitation trajectories. Values are means across seeds 42, 43, and 44; the selected setting ($k=7$) is \underline{underlined}. Model abbreviations follow Table~\ref{tab:value_scores_args_observers}. Standard errors are at most .0045 for WA-based measures and .464 topics for Distinct Topics.}\label{tab:k_sensitivity_analysis}
    \footnotesize
    
    \begin{tabular}{
    m{0.44cm}
    >{\centering\arraybackslash}m{0.45cm} %Selected WA
    >{\centering\arraybackslash}m{0.45cm}
    >{\centering\arraybackslash}m{0.45cm}|
    >{\centering\arraybackslash}m{0.45cm} %Candidate WA
    >{\centering\arraybackslash}m{0.45cm}
    >{\centering\arraybackslash}m{0.45cm}|
    >{\centering\arraybackslash}m{0.45cm} %Candidate Max WA				
    >{\centering\arraybackslash}m{0.45cm}
    >{\centering\arraybackslash}m{0.45cm}|
    >{\centering\arraybackslash}m{0.45cm} %Selects Max WA
    >{\centering\arraybackslash}m{0.45cm}
    >{\centering\arraybackslash}m{0.45cm}|
    >{\centering\arraybackslash}m{0.45cm} %Distinct Topics		
    >{\centering\arraybackslash}m{0.45cm}
    >{\centering\arraybackslash}m{0.45cm}
}
    \toprule
    Metric $\rightarrow$ & \multicolumn{3}{c}{\makecell{Selected\\WA}} & \multicolumn{3}{c}{\makecell{Candidate\\WA}} & \multicolumn{3}{c}{\makecell{Candidate\\Max WA}} & \multicolumn{3}{c}{\makecell{Selects\\Max WA}} & \multicolumn{3}{c}{\makecell{Distinct\\Topics}}  \\
    \cmidrule(lr){2-4} \cmidrule(lr){5-7} \cmidrule(lr){8-10} \cmidrule(lr){11-13} \cmidrule(lr){14-16} 
    k $\downarrow$ & L8 & Q14 & M8 & L8 & Q14 & M8 & L8 & Q14 & M8 & L8 & Q14 & M8 & L8 & Q14 & M8 \\
    
    \midrule

    3  & .838 & .825 & .824 & .792 & .791 & .791 & .936 & .935 & .935 & .472 & .442 & .438 & 29.7 & 35.4 & 26.7\\
    5  & .853 & .832 & .831 & .793 & .790 & .789 & .970 & .970 & .969 & .369 & .327 & .325 & 34.4 & 41.2 & 31.2\\
    \underline{7}  & \underline{.860} & \underline{.840} & \underline{.837} & \underline{.793} & \underline{.791} & \underline{.790} & \underline{.984} & \underline{.984} & \underline{.983} & \underline{.329} & \underline{.292} & \underline{.290} & \underline{36.5} & \underline{44.7} & \underline{34.0}\\
    9  & .866 & .843 & .841 & .792 & .791 & .789 & .991 & .991 & .991 & .323 & .279 & .275 & 37.8 & 46.0 & 34.3\\
    11 & .868 & .846 & .843 & .793 & .792 & .790 & .995 & .995 & .995 & .312 & .271 & .265 & 38.9 & 47.1 & 35.3\\
    13 & .873 & .847 & .843 & .792 & .792 & .789 & .997 & .997 & .996 & .316 & .271 & .257 & 40.0 & 48.2 & 35.2\\
    
    \bottomrule
    \end{tabular}
    \vspace{-10pt}

\end{table}

%% ----------------------------------------------------- %%
\vspace{-5pt}
\subsection{Breadth--Depth Behavior}
%\vspace{-5pt}
%% ----------------------------------------------------- %%

%% ----------------------------------------------------- %%
\vspace{-5pt}
\subsubsection{Differences Across Models}
%% ----------------------------------------------------- %%

We map the information-seeking preferences previously characterized onto the breadth--depth behavior they induce over sequential elicitation, without explicitly specifying or optimizing for breadth, depth, topic coverage, or switching. We hold the elicitation environment and selection rule fixed and vary only the LLM used to assign information-seeking scores. We characterize each trajectory by its topic-switch rate, number of distinct selected topics, and mean length of consecutive same-topic runs, and report each metric averaged across trajectories and seeds. Table~\ref{tab:breadth_depth_model_set} shows substantial variation in the resulting trajectories: topic-switch rates range from .295 (P4) to .560 (Q14), with distinct-topic counts ranging from 23.4 to 44.7 and mean run lengths from 1.76 to 3.35. Most models also select higher-WA arguments than random despite argument quality not appearing in the elicitation objective. Thus, differences in local information-seeking preferences are sufficient to induce markedly different breadth--depth behavior over the course of elicitation.

%% ----------------------------------------------------- %%
%\vspace{-10pt}
\subsubsection{Effect of Available Opportunities}
\label{sec:available_opportunities}
%% ----------------------------------------------------- %%

\begin{wraptable}[16]{r}{0.47\linewidth}

    \vspace{-40pt}
    \centering

    \caption{Scores by history, semantic similarity, and prior confirmation (three-seed means). Conf.: confirmed; Not Conf.: unconfirmed. \textbf{Bold} indicates contrasts discussed in text.}\label{tab:similarity_groups}
    \footnotesize
    
    \begin{tabular}{
    m{0.6cm}
    >{\centering\arraybackslash}m{.8cm}
    >{\centering\arraybackslash}m{.6cm}
    >{\centering\arraybackslash}m{.6cm}|
    >{\centering\arraybackslash}m{.6cm}
    >{\centering\arraybackslash}m{.6cm}
    >{\centering\arraybackslash}m{.6cm}
}
    \toprule
    & & \multicolumn{2}{c}{\makecell{W/ History}} & \multicolumn{2}{c}{\makecell{W/out History}}  \\
    \cmidrule(lr){3-4} \cmidrule(lr){5-6}
    Model & Group & Conf. & Not Conf. & Conf. & Not Conf. \\
    
    \midrule

    \multirow[c]{2}{*}{Q7}

    & $\leq0.7$	& .471 & .485 & .670 & .667\\
    & $>0.7$	& .961 & .945 & .952 & .946 \\

    \multirow[c]{2}{*}{Q14}
	
    & $\leq0.7$	& .430 & .481 & .622 & .619 \\
    & $>0.7$	& \textbf{.257} & .649 & \textbf{.823} & .833 \\

    \multirow[c]{2}{*}{M8}
					
    & $\leq0.7$	& .663 & .677 & .598 & .594 \\
    & $>0.7$	& .736 & .739 & .631 & .635 \\

    \multirow[c]{2}{*}{M14}
				
    & $\leq0.7$	& .724 & .739 & .837 & .838 \\
    & $>0.7$	& \textbf{.634} & .718 & \textbf{.865} & .864 \\

    \multirow[c]{2}{*}{M24}
					
    & $\leq0.7$	& .831 & .861 & .920 & .921 \\
    & $>0.7$	& \textbf{.708} & .844 & \textbf{.931} & .934 \\

    \bottomrule
    \end{tabular}

\end{wraptable}

The preceding analysis holds the candidate-set size fixed, but realized breadth and depth may also depend on the opportunities available to the elicitor. We therefore vary $k\in\{3,5,7,9,11,13\}$, changing the number of hypotheses available at each step while holding the remaining simulation parameters fixed. To characterize the hypotheses available and selected as the candidate set changes using an external human-annotated measure, we use the Weighted-Average (WA) argument-quality scores provided by IBM-Rank-30k~\citep{gretz_large-scale_2019}. WA aggregates human judgments of whether an argument should be used as-is in a speech on its topic, weighted by annotator reliability. We use WA only as an external, human-annotated property of the underlying arguments; it is not intended as a measure of information-seeking preference or as the objective optimized by the elicitor. We report this analysis for representative models from three model families.

Table~\ref{tab:k_sensitivity_analysis} shows that realized information-seeking behavior depends on the opportunities available to the elicitor. Increasing $k$ from 3 to 13 increases the number of distinct topics explored for all three models, from 29.7 to 40.0 for L8, 35.4 to 48.2 for Q14, and 26.7 to 35.2 for M8. The ordering across models remains consistent, with Q14 exploring the most topics and M8 the fewest, showing that candidate-set size shifts elicitation breadth without eliminating differences induced by model-specific information-seeking preferences. Larger candidate sets improve the best available argument with rapidly diminishing returns: maximum candidate WA rises from approximately .935 at $k=3$ to .984 at $k=7$ and .997 at $k=13$, while selected WA increases more gradually and the maximum-WA selection rate generally decreases with $k$. We therefore use $k=7$ when candidate-set size is held fixed, balancing available opportunities against the cost of scoring additional candidates. Together, these patterns characterize $k$ as an opportunity constraint: increasing it expands both the range and quality of available information, while the model's preferences continue to determine which opportunities are pursued.

%% ----------------------------------------------------- %%
\vspace{-7pt}
\subsection{History-Dependent Information-Seeking Preferences}
\label{sec:history_changes_value}
%% ----------------------------------------------------- %%

%% ----------------------------------------------------- %%
%\vspace{-5pt}
\subsubsection{Redundancy and Confirmation}
%% ----------------------------------------------------- %%

Information-seeking preferences may depend on what has already been learned. We examine this dependence in the case of potentially redundant information by providing the model with the topic-specific interaction history and explicitly instructing it that repeated information is less valuable, while leaving the model to determine what constitutes repetition and how strongly it should affect the assessment. This explicit instruction is important: as we show in Section~\ref{sec:neutral_prompt}, providing interaction history alone does not reliably cause models to discount repeated information. No separate deduplication procedure, similarity threshold, redundancy penalty, or candidate-filtering rule is used in selection; any effect of redundancy instead operates through the same information-seeking score used for all candidates. To isolate the effect of interaction history, at each elicitation step we score the same candidate hypotheses both with and without the topic-specific history.

For this analysis, we measure the semantic similarity of candidate hypotheses to previously selected hypotheses using \textit{cross-encoder/stsb-roberta-large} \citep{reimers_sentence-bert_2019}, and group candidates by similarity ($>0.7$ vs. $\leq0.7$); examples of highly similar candidate--history pairs are provided in Appendix~\ref{sec:similarity_example}. We further distinguish whether the similar information was previously confirmed or only unconfirmed, which allows us to test whether a candidate's score depends on its semantic relationship to earlier hypotheses and their confirmation status. Table~\ref{tab:similarity_groups} shows that history can substantially change how models evaluate potentially redundant information. For highly similar candidates whose overlapping information was previously confirmed, including history reduces mean scores from .823 to .257 for Q14, .865 to .634 for M14, and .931 to .708 for M24. Model size also plays a role, with this distinction absent for Q7 and M8 but emerging for Q14 and the larger M14 and M24 models. Moreover, the reduction is stronger when the overlapping information was previously confirmed than when it was unconfirmed, indicating that these models are more willing to move on once similar information has already been confirmed.

%% ----------------------------------------------------- %%
%\vspace{-10pt}
\subsubsection{Effect on Subsequent Selection}
%% ----------------------------------------------------- %%

\begin{wraptable}[15]{r}{0.4\linewidth}

    \vspace{-45pt}
    \centering

    \caption{Selection rates for candidates similar to confirmed (StC) or only to unconfirmed (SOtU) information, with and without history (three-seed means).}\label{tab:confirmation_flipping}
    \footnotesize
    
    \begin{tabular}{
    m{.7cm}
    >{\centering\arraybackslash}m{0.7cm}
    >{\centering\arraybackslash}m{0.7cm}|
    >{\centering\arraybackslash}m{0.7cm}
    >{\centering\arraybackslash}m{0.7cm}
    }
    
    \toprule
    & \multicolumn{2}{c}{\makecell{W/ History}} & \multicolumn{2}{c}{\makecell{W/out History}}  \\
    \cmidrule(lr){2-3} \cmidrule(lr){4-5}
    Model & StC & SOtU & StC & SOtU \\
    
    \midrule

    L3	& .315	& .276 & .330	& .301 \\
    L8	& .201	& .216 & .251	& .238 \\
    L70	& \textbf{.033}  &.064  & \textbf{.216}   & .211 \\
    Q7	& .287	& .233 & .280	& .265 \\
    Q14	& \textbf{.042}	& .120 & \textbf{.211}	& .224 \\
    %Q72	&       &      &        &      \\
    M8	& .230	& .172 & .184	& .184 \\
    M14	& \textbf{.053}	& .090 & \textbf{.180}	& .193 \\
    M24	& \textbf{.049}	& .092 & \textbf{.167}	& .163 \\
    P4	& .467	& .354 & .344	& .310 \\
    G12	& .066	& .124 & .113	& .118 \\

    \bottomrule
    \end{tabular}

\end{wraptable}

We next examine whether these changes in information-seeking scores propagate to the elicitor's subsequent selections. Because the elicitor selects the hypothesis with the highest information-seeking score from the candidate set, a candidate's score need not decrease substantially to change the selection; it need only fall below that of another available candidate. This is a useful property of the elicitation process, as it allows redundant information to fall out of selection without requiring a separate mechanism for deduplication, removal, or filtering. At each trajectory step, we classify each candidate according to whether it is similar to previously confirmed information or only to previously unconfirmed information. We then compute each group's selection rate as the proportion of its candidates that are selected.

Table~\ref{tab:confirmation_flipping} shows that the effect of history on subsequent selection also varies with model size. Within Qwen, Q7 shows little change in the selection of candidates similar to confirmed information (.280 without history vs. .287 with history), whereas Q14 shows a large reduction (.211 to .042). A similar pattern appears within Mistral: M8 shows no reduction (.184 to .230), while M14 and M24 fall from .180 to .053 and from .167 to .049, respectively. The Llama models show a related but less monotonic pattern, with a large reduction for L70 (.216 to .033) but smaller reductions for L3 and L8. For Q14, M14, and M24, the reduction is also larger for candidates similar to confirmed information than for those similar only to unconfirmed information. These results show that the history-dependent differences observed in information-seeking scores propagate to subsequent selection, with the clearest effects generally appearing in larger models. When overlap with previously confirmed information lowers a candidate's score, that candidate becomes less likely to be selected, allowing redundancy to affect selection without explicit deduplication or filtering.

%% ----------------------------------------------------- %%
\vspace{-5pt}
\subsection{Sensitivity and Ablations}
%% ----------------------------------------------------- %%

%% ----------------------------------------------------- %%
\subsubsection{Label Sensitivity}
\label{sec:label_validation}
%% ----------------------------------------------------- %%

\begin{wrapfigure}[9]{r}{0.56\linewidth}
    \vspace{-50pt}
    \centering
    \includegraphics[height=3.6cm]{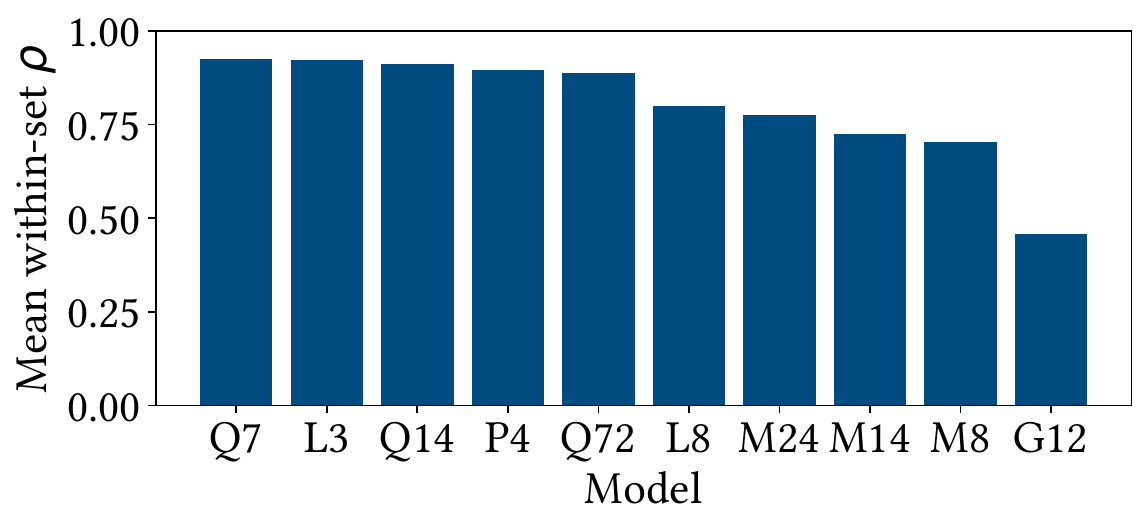}
    \vspace{-20pt}
    \caption{Response-label formulation robustness. Mean within-set rank correlation between \textit{Yes} and \textit{Valuable}.}\label{fig:label_correlations_analysis}
\end{wrapfigure}

To assess sensitivity to the response-label interface, we rescored every candidate encountered along the original trajectories using an alternative response set, replacing \textit{Yes}/\textit{No}/\textit{Uncertain} with \textit{Valuable}/\textit{Not valuable}/\textit{Uncertain}, while holding the candidate, objective, and interaction history fixed. We compute the Spearman correlation between the rankings induced by the two affirmative probabilities within each candidate set and average these correlations across decision points. Figure~\ref{fig:label_correlations_analysis} shows substantial agreement for most models: nine of ten have mean correlations above .70, including five above .88, while G12 is substantially lower at .459. The positive correlations across all models indicate that the induced information-seeking preferences are not specific to the \textit{Yes} formulation, although the degree of robustness varies across models. Additional analysis of the impact of these differences in trajectories is available in Appendix~\ref{sec:label_trajectory_sensitivity}.

%% ----------------------------------------------------- %%
\subsubsection{Prompt Sensitivity}
\label{sec:neutral_prompt}
%% ----------------------------------------------------- %%

\begin{wraptable}[15]{r}{0.45\linewidth}

    \vspace{-45pt}
    \centering

    \caption{Scores for highly similar candidates ($>0.7$) under the original and neutral prompts (three-seed means).}\label{tab:prompt_sensitivity}
    \footnotesize
    
    \begin{tabular}{
    m{0.7cm}
    >{\centering\arraybackslash}m{.8cm}
    >{\centering\arraybackslash}m{.6cm}
    >{\centering\arraybackslash}m{.6cm}|
    >{\centering\arraybackslash}m{.6cm}
    >{\centering\arraybackslash}m{.6cm}
    >{\centering\arraybackslash}m{.6cm}
}
    \toprule
    & & \multicolumn{2}{c}{\makecell{W/ History}} & \multicolumn{2}{c}{\makecell{W/out History}}  \\
    \cmidrule(lr){3-4} \cmidrule(lr){5-6}
    Model & Prompt & Conf. & Not Conf. & Conf. & Not Conf. \\
    
    \midrule

    \multirow[c]{2}{*}{Q7}

    & Original	& .961 & .945 & .952 & .946 \\
    & Neutral	& .979 & .968 & .957 & .958 \\

    \multirow[c]{2}{*}{Q14}
	
    & Original	& \textbf{.257} & .649 & \textbf{.823} & .833 \\
    & Neutral	& \textbf{.887} & .929 & \textbf{.899} & .925 \\

    \multirow[c]{2}{*}{M8}
					
    & Original	& .736 & .739 & .631 & .635 \\
    & Neutral	& .947 & .944 & .940 & .945 \\
    
    \multirow[c]{2}{*}{M14}
				
    & Original	& \textbf{.634} & .718 & \textbf{.865} & .864 \\
    & Neutral	& \textbf{.915} & .969 & \textbf{.877} & .969  \\
    
    \multirow[c]{2}{*}{M24}
					
    & Original	& \textbf{.708} & .844 & \textbf{.931} & .934 \\
    & Neutral	& \textbf{.962} & .987 & \textbf{.950} & .986 \\
    
    \bottomrule
    \end{tabular}

\end{wraptable}

We test whether interaction history by itself is sufficient to produce the redundancy-sensitive behavior above by replacing the original prompt with a neutral variant that does not explicitly ask the model to consider repetition or additional information. As shown in Table~\ref{tab:prompt_sensitivity}, the reductions in information-seeking scores for hypotheses similar to previously confirmed information largely disappear under the neutral prompt, most notably for Q14, M14, and M24. History conditioning alone therefore does not reliably produce redundancy-sensitive assessments. Instead, the results show that redundancy sensitivity depends on how the information-seeking judgment is specified: when redundancy is made relevant to the assessment, it can affect candidate scores and subsequent selections through the same scoring mechanism used throughout the simulation.

%% -- ################################################## %%
\vspace{-10pt}
\section{Conclusion}
%% -- ################################################## %%

We studied how model-dependent information-seeking preferences shape sequential elicitation. Our simulation intentionally abstracts away question generation, natural-language responses, and an unbounded information space, allowing us to isolate the effects of LLM information-seeking preferences from other components of dialogue at scale. Extending this analysis to less controlled settings with richer interaction dynamics is an important direction for future work. We also treat the elicitor as the observer assessing which information is worth pursuing, although information may ultimately be valued by other observers. For example, an interviewer's judgments may depend on what they expect an audience to find valuable. Studying how information-seeking preferences interact across elicitors, audiences, and other observers therefore provides another direction for future work.

\bibliography{references}
\bibliographystyle{iclr2027_conference}

\newpage

\appendix

%change figure numbering
\renewcommand{\thefigure}{\Alph{section}.\arabic{figure}}
\renewcommand{\thetable}{\Alph{section}.\arabic{table}}
\setcounter{figure}{0}
\setcounter{table}{0}

%% -- ################################################## %%
\section{Prompts}
\label{sec:appendix_prompts}
%% -- ################################################## %%

The following prompt was used for the main experiments.

\begin{tcolorbox}[
    title={Main Prompt},
    colback=gray!5,
    colframe=gray!50,
    boxrule=0.5pt,
    arc=2pt
]
\ttfamily\fontsize{7.5}{8}\selectfont
You are interviewing a person. Your objective is to \{dataset\_objective\} and the current topic is '\{topic\}'.

\medskip

The following is information you have already gathered about this topic:
\{history\}

\medskip

Regardless of whether you agree with what they say, does the response provide valuable information about the person's views? Repeated information makes it less valuable.

\medskip

Respond with exactly one of the following:

\medskip

\{valid\_labels\}

\end{tcolorbox}

For the prompt-sensitivity ablation in Section~\ref{sec:neutral_prompt}, we remove the instruction that repeated information is less valuable.

\begin{tcolorbox}[
    title={Ablation Prompt},
    colback=gray!5,
    colframe=gray!50,
    boxrule=0.5pt,
    arc=2pt
]
\ttfamily\fontsize{7.5}{8}\selectfont
You are interviewing a person. Your objective is to \{dataset\_objective\} and the current topic is '\{topic\}'.

\medskip

The following is information you have already gathered about this topic:
\{history\}

\medskip

Regardless of whether you agree with what they say, does the response provide valuable information about the person's views?

\medskip

Respond with exactly one of the following:

\medskip

\{valid\_labels\}

\end{tcolorbox}

%% -- ################################################## %%
\section{With and Without Uncertain Label}
\label{sec:with_without_uncertain}
%% -- ################################################## %%

Table~\ref{tab:with_without_uncertain} compares assessments of the same argument--objective pairs with and without \textit{Uncertain} as a response option. Although little probability is assigned to \textit{Uncertain} itself, its inclusion substantially affects $P(\textit{Yes})$. Without \textit{Uncertain}, intermediate scores largely shift toward the extremes, while high scores remain comparatively stable. Including \textit{Uncertain} therefore preserves greater differentiation among intermediate information-seeking scores.

\begin{table}

    %\vspace{-20pt}
    \centering

    \caption{Effect of including \textit{Uncertain} on information-seeking scores. $\text{Yes}\Delta$ denotes the change in $P(\textit{Yes})$ when \textit{Uncertain} is included.}\label{tab:with_without_uncertain}
    \footnotesize
    
    \begin{tabular}{
    m{6.5cm}
    >{\centering\arraybackslash}m{0.73cm}
    >{\centering\arraybackslash}m{0.73cm}
    >{\centering\arraybackslash}m{0.73cm}|
    >{\centering\arraybackslash}m{0.73cm}
    >{\centering\arraybackslash}m{0.73cm}|
    >{\centering\arraybackslash}m{0.73cm}
}
    \toprule
    & \multicolumn{3}{c}{\makecell{W/ Uncert.}} & \multicolumn{2}{c}{\makecell{W/out Uncert.}} & \\
    \cmidrule(lr){2-4} \cmidrule(lr){5-6}
    &Yes& No& Uncert.& Yes& No& Yes$\Delta$ \\
    
    \midrule

    We should adopt a multi-party system                    & 0.94 & 0.05 & 0.02 & 0.93 & 0.08 & 0.01\\
    We should introduce compulsory voting                   & 0.86 & 0.13 & 0.02 & 0.79 & 0.21 & 0.07\\
    We should oppose collectivism                          	& 0.82 & 0.14 & 0.04 & 0.75 & 0.25 & 0.07\\
    We should adopt libertarianism                        	& 0.79 & 0.15 & 0.06 & 0.75 & 0.25 & 0.05\\
    We should adopt an austerity regime                  	& 0.71 & 0.25 & 0.04 & 0.50 & 0.50 & 0.21\\
    We should limit judicial activism                      	& 0.63 & 0.34 & 0.03 & 0.36 & 0.64 & 0.27\\
    We should end the use of economic sanctions         	& 0.62 & 0.36 & 0.02 & 0.36 & 0.64 & 0.26\\
    Homeopathy brings more harm than good               	& 0.53 & 0.46 & 0.01 & 0.17 & 0.83 & 0.36\\
    Blockade of the Gaza Strip should be ended          	& 0.48 & 0.51 & 0.01 & 0.21 & 0.79 & 0.27\\
    Entrapment should be legalized                        	& 0.47 & 0.51 & 0.02 & 0.18 & 0.82 & 0.29\\
    We should abolish the right to keep and bear arms   	& 0.46 & 0.53 & 0.01 & 0.18 & 0.82 & 0.28\\
    Intelligence tests bring more harm than good        	& 0.45 & 0.54 & 0.01 & 0.10 & 0.90 & 0.35\\
    We should fight urbanization                        	& 0.42 & 0.56 & 0.02 & 0.14 & 0.86 & 0.27\\
    Holocaust denial should be a criminal offence       	& 0.41 & 0.58 & 0.02 & 0.14 & 0.87 & 0.27\\
    We should fight for the abolition of nuclear weapons	& 0.41 & 0.58 & 0.02 & 0.14 & 0.86 & 0.26\\

    \bottomrule
    \end{tabular}

\end{table}

%% -- ################################################## %%
\section{Label Trajectory Sensitivity}
\label{sec:label_trajectory_sensitivity}
%% -- ################################################## %%

We repeat the sequential elicitation experiment using the alternative labels \texttt{Valuable}, \texttt{Not valuable}, and \texttt{Uncertain}, while holding the assessment question, simulation environment, and selection rule fixed. Table~\ref{tab:label_val} compares breadth--depth behavior across independently generated trajectories under the two label formulations. Sensitivity varies across models: Q7, L3, and M24 exhibit similar behavior under both formulations, whereas M8 changes substantially. These results show that differences between the two label formulations can propagate to trajectory-level behavior, with the magnitude of the effect varying substantially across models.

\begin{table}

    \vspace{-20pt}
    \centering

    \caption{Breadth--depth behavior under the original \textit{Yes}/\textit{No}/\textit{Uncertain} labels (Yes) and alternative \textit{Valuable}/\textit{Not valuable}/\textit{Uncertain} labels (Valuable) (three-seed means).}\label{tab:label_val}
    \footnotesize
    
    \begin{tabular}{
        m{0.8cm}
        >{\centering\arraybackslash}m{1.cm}
        >{\centering\arraybackslash}m{1.cm}
        >{\centering\arraybackslash}m{1.cm}
        >{\centering\arraybackslash}m{1.cm}
        >{\centering\arraybackslash}m{1.cm}
        >{\centering\arraybackslash}m{1.cm}
        >{\centering\arraybackslash}m{1.cm}
        >{\centering\arraybackslash}m{1.cm}
        >{\centering\arraybackslash}m{1.cm}
    }
    \toprule
    
    & \multicolumn{3}{c}{Topic Switch Rate}
    & \multicolumn{3}{c}{Distinct Topics}
    & \multicolumn{3}{c}{Topic Run Length} \\
    \cmidrule(lr){2-4}\cmidrule(lr){5-7}\cmidrule(lr){8-10}
    Model & Yes & Valuable & $|\Delta|$ & Yes & Valuable & $|\Delta|$  & Yes & Valuable & $|\Delta|$  \\
    \midrule
   
    L3	& .348 & .331 & 0.017 & 27.177 & 25.977 & 1.200	 & 2.834	 & 2.986	 & 0.152 \\
    L8	& .458 & .561 & 0.103 & 36.523 & 44.907 & 8.384	 & 2.158	 & 1.760	 & 0.398 \\
    Q7	& .551 & .555 & 0.004 & 43.810 & 43.767 & 0.043	 & 1.794	 & 1.781	 & 0.013 \\
    Q14	& .560 & .521 & 0.039 & 44.720 & 41.680 & 3.040	 & 1.764	 & 1.897	 & 0.133 \\
    M8	& .413 & .666 & 0.253 & 33.997 & 54.230 & 20.233 & 2.542     & 	1.485    & 1.057 \\
    M14	& .526 & .449 & 0.077 & 42.657 & 36.713 & 5.944	 & 1.878	 & 2.198	 & 0.320 \\
    M24	& .484 & .466 & 0.018 & 38.977 & 37.997 & 0.980	 & 2.040	 & 2.118	 & 0.078 \\
    
    \bottomrule
    \end{tabular}
\end{table}

%% -- ################################################## %%
\section{Semantic Similarity Examples}
\label{sec:similarity_example}
%% -- ################################################## %%

Table~\ref{tab:similarity_example} shows that the information space contains many highly similar candidate--history pairs, with similarity scores reaching .97. Despite this redundancy, overlapping information can be deprioritized through the information-seeking process itself, without explicit deduplication, removal, or candidate filtering.

\begin{table}
    \vspace{-20pt}
    \centering
    \footnotesize
    \caption{Ten highest-similarity pairs from Qwen2.5-14B-Instruct trajectories.}\label{tab:similarity_example}

    \renewcommand{\arraystretch}{1.25}
    \begin{tabular}{
        >{\raggedright\arraybackslash}m{6.1cm}
        >{\raggedright\arraybackslash}m{6.1cm}
        >{\centering\arraybackslash}m{.5cm}
    }
    \toprule

        Candidate & History Argument & Sim. Score\\
        \midrule

    The vow of celibacy should be abandoned because it's outdated in today's society. & The vow of celibacy is outdated in today's society and so it should be abandoned.& 0.97 \\
    We should not ban telemarketing because it would put people out of jobs. & We should not ban telemarketing because it would result in a loss of jobs.& 0.97 \\
    Burning the flag is freedom of speech and should be allowed. & Flag burning is a form of free speech and should not be banned. & 0.96 \\
    We should abolish the right to keep and bear arms because there are those who abuse that power and go on rampages killing innocent people and shooting up schools. & We should abolish the right to keep and bear arms because too many abuse that right by killing other and going on rampages and school shootings. & 0.95 \\
    If we legalise the organ trade then it will mean that more lives can be saved. & Legalizing organ trade has the potential to save many lives. & 0.95 \\
    People should be allowed to work for as long as they want and as long as they are capable. & Individuals should be able to work as long as they are willing and able. & 0.95 \\
    Prohibiting school prayer is the right thing to do to avoid conflict in the classroom. & School prayer promotes conflict in the classroom and therefore should be prohibited. & 0.94 \\
    Safe spaces tend to act as a brake on free speech. & Safe spaces are often used to stifle free expression of opinions. & 0.94 \\
    Students are children who need room to learn and grow from mistakes. & Students are still children who need room to learn and grow from mistakes.& 0.94 \\
    Social media can be used as a bully technique.& Social media is an outlet for bullying.& 0.94 \\

    \bottomrule
    \end{tabular}
\end{table}

\end{document}